\documentclass[runningheads]{llncs}
\usepackage[T1]{fontenc}
\usepackage{graphicx}
\usepackage{latexsym}
\usepackage{amsmath}
\usepackage{mathtools}
\usepackage{amssymb}
\usepackage{comment}
\usepackage{stmaryrd}
\usepackage{algorithm}
\usepackage{algpseudocode}
\usepackage{caption} 
\usepackage[title]{appendix}
\begin{document}

\newcommand{\alg}[1]{\mathsf{#1}}
\newcommand{\Prover}{\alg{P}}
\newcommand{\Verifier}{\alg{V}}
\newcommand{\Simulator}{\alg{S}}
\newcommand{\PPT}{\alg{PPT}}
\newcommand{\isom}{\cong}
\newcommand{\from}{\stackrel{\scriptstyle R}{\leftarrow}}
\newcommand{\handout}[5]{
	\noindent
	\begin{center}
		\framebox{
			\vbox{
				\hbox to 5.78in { {\bf Hybrid Systems} \hfill #2 }
				\vspace{4mm}
				\hbox to 5.78in { {\Large \hfill #5  \hfill} }
				\vspace{2mm}
				\hbox to 5.78in { {\it #3 \hfill #4} }
			}
		}
	\end{center}
	\vspace*{4mm}
}

\newcommand{\ho}[5]{\handout{#1}{#2}{Guide:
		#3}{#4}{#5}}
\newcommand{\al}{\alpha}
\newcommand{\nd}{\wedge}
\newcommand{\defn}{\coloneqq}
\newcommand{\Z}{\mathbb Z}
\newcommand{\real}{\mathbb{R}}
\newcommand{\nat}{\mathbb{N}}
\newcommand{\calH}{\mathcal{H}}
\newcommand{\ppcd}{\text{PPCD}}
\newcommand{\reduce}[1]{#1^{\textit{red}}}
\newcommand{\loc}{Q}
\newcommand{\state}{\mathcal{X}}
\newcommand{\poly}[1]{Poly(#1)}
\newcommand{\inv}{\textit{Inv}}
\newcommand{\flow}{\textit{Flow}}
\newcommand{\guard}{\textit{Guard}}
\newcommand{\eguard}{\mathcal{G}}
\newcommand{\edges}{\textit{Edges}}
\newcommand{\dist}[1]{Dist(#1)}
\newcommand{\itstar}{\item[$\bigstar$]}
\newcommand{\vertx}{V}
\newcommand{\gredge}{E}
\newcommand{\ewt}{W}
\newcommand{\ec}{E_c}
\newcommand{\ep}{E_p}
\newcommand{\graph}{G}
\newcommand{\initnode}{I_0}
\newcommand{\initst}{s_{\textit{init}}}
\newcommand{\stablest}{s_0}
\newcommand{\pf}{P} 
\newcommand{\Path}{\sigma}
\newcommand{\infpath}{\sigma_{\infty}}
\newcommand{\len}[1]{len(#1)}
\newcommand{\indx}{I}
\newcommand{\union}{\cup}
\newcommand{\bigunion}{\bigcup}
\newcommand{\intersect}{\cap}
\newcommand{\bigintersect}{\bigcap}
\newcommand{\ball}[2]{B_{#1}(#2)}
\newcommand{\prob}{\textit{Pr}}
\newcommand{\probpath}[2]{P_{#1}(#2)}
\newcommand{\tildeprob}{\tilde{\textit{Pr}}}
\newcommand{\fpath}[1]{\textit{Paths}_{\textit{fin}}(#1)}
\newcommand{\cyl}[1]{\textit{Cyl}(#1)}
\newcommand{\apath}[1]{\textit{Paths}(#1)}
\newcommand{\ipath}[1]{\textit{Paths}(#1)}
\newcommand{\spath}{\mathcal{SP}}
\newcommand{\scyl}{\mathcal{SC}}
\newcommand{\edgeset}{\mathcal{E}}
\newcommand{\sigal}{\mathcal{F}}
\newcommand{\states}{S}
\newcommand{\ctran}{\stackrel{c}{\rightarrow}}
\newcommand{\ptran}{\stackrel{p}{\rightarrow}}
\newcommand{\tran}{\text{P}}
\newcommand{\stran}{S_\rightarrow}
\newcommand{\sstran}[1]{{#1}_{\rightarrow}}
\newcommand{\metric}{d}
\newcommand{\weight}{\text{W}}
\newcommand{\calM}{\mathcal{M}}
\newcommand{\dtmc}{\text{DTMC}}
\newcommand{\restrict}[2]{#1|#2}
\newcommand{\calMw}{\mathcal{M}_W}
\newcommand{\wdtmc}{\text{WDTMC}}
\newcommand{\boundary}[1]{\partial(#1)}
\newcommand{\norm}[2]{\lvert\!\lvert #1 \rvert\!\rvert}
\newcommand{\pathset}[3]{\Sigma_{#1}^{#2}#3}
\newcommand{\partition}{\mathcal{P}}
\newcommand{\size}[1]{\vert #1\vert}
\newcommand{\semantics}[1]{\mathcal{M}_{#1}}
\newcommand{\abtrct}[2]{#1/#2}
\newcommand{\face}{\mathbb{F}}
\newcommand{\cycle}{\mathcal{C}}
\newcommand{\wld}{S_\sigma}
\newcommand{\ldecomp}[1]{#1^{\mathcal{L}}}
\newcommand{\fdecomp}[1]{#1^d}
\newcommand{\pcdecomp}[1]{#1^{\mathcal{SP}\union\mathcal{SC}}}
\newcommand{\tildewt}{\tilde{W}}
\newcommand{\rhost}{\rho^*}
\newcommand{\pst}[1]{p^{\textit{st}}(#1)}
\newcommand{\mstep}[1]{\overset{\mathrm{#1}}{\leadsto}}
\newcommand{\cov}{\textit{Cov}}
\newcommand{\var}{\textit{Var}}
\newcommand{\linineq}[1]{L_{#1}}
\newcommand{\coreq}[1]{{#1}_{\textit{eq}}}
\newcommand{\disunion}{\sqcup}
\newcommand{\pathprop}{\mathbb{P}}

\title{Stability Analysis of Planar Probabilistic Piecewise Constant Derivative Systems\thanks{This work was 
partially supported by NSF CAREER Grant No. 1552668 and NSF Grant No. 2008957.}}
\titlerunning{Stability of $\ppcd$}
%
\author{Spandan Das\inst{1}\orcidID{0000-0002-1995-2592} \and
Pavithra Prabhakar\inst{1}}
\authorrunning{S. Das and P. Prabhakar}
%
\institute{Kansas State University, Manhattan, USA \\
\email{\{spandan,pprabhakar\}@ksu.edu}
}
\maketitle              
\begin{abstract}
In this paper, we study the probabilistic stability analysis of a
subclass of stochastic hybrid systems, called the \emph{Planar Probabilistic
	Piecewise Constant Derivative Systems (Planar $\ppcd$)}, where the continuous dynamics
is deterministic, constant rate and planar, the discrete switching
between the modes is probabilistic and happens at boundary of the
invariant regions, and the continuous states are not reset during
switching. These aptly model piecewise linear behaviors of planar
robots.
Our main result is an exact algorithm for deciding \emph{absolute} and
\emph{almost sure stability} of Planar $\ppcd$
under some mild assumptions on mutual reachability between the states
and the presence of non-zero probability self-loops.
Our main idea is to reduce the stability problems on planar $\ppcd$ into
corresponding problems on Discrete-time Markov Chains with edge weights.

\end{abstract}
\section{Introduction}
Stability of Stochastic Hybrid Systems (SHS)~\cite{rutten2004mathematical} is a desirable property, as it
guarantees eventual convergence of executions to a point of equilibrium, even in the presence of random errors.
In this paper, we investigate the stability of a certain kind of SHS where the continuous state space is planar and dynamics has constant rate, where the rates are discrete and chosen probabilistically.
More precisely, we study \emph{Probabilistic Piecewise Constant
  Derivative Systems ($\ppcd$)}, that consist of a finite number of
discrete states representing different modes of operation each
associated with a constant rate dynamics, and probabilistic mode
switches enabled at certain polyhedral boundaries. 
Such systems can aptly model piecewise linear behaviour of planar robots.

Safety analysis of SHS has been extensively studied in the context of
both non-stochastic as well as stochastic hybrid 
systems~\cite{prajna2004safety,lal2018hierarchical,clarke2003abstraction,alur2003counter,lal2019counterexample};
stability on the other hand is relatively less explored, especially,
from a computational point of view.
It is well-known that even for non-stochastic hybrid systems
decidability (existence of exact algorithms) for safety is achievable
only under restrictions on the dynamics and the dimension~\cite{HENZINGER199894}.
More recently, decidability of stability of hybrid systems has been
explored in the non-stochastic setting~\cite{prabhakar2013decidability}.
The main contribution of this paper is the identification of a
practically useful subclass of stochastic hybrid systems for which
stability is decidable along with an exact stability analysis algorithm.

The classical stability analysis techniques build on the notion of
Lyapunov functions that provide a certificate of stability.
While the notion of Lyapunov functions have been extended to the
hybrid system setting, computing them is a challenge.
Typically, they require solving certain complex optimization problems,
for instance, to deduce coefficients of polynomial templates, and more
importantly, need the exploration of increasingly complex templates.
In this paper, we take an alternate route where we present graph
theory based reductions to show the decidability of stability analysis.

Our broad approach is to reduce a planar $\ppcd$, that is a potentially
infinite state probabilistic system, to that of a Finite State
Discrete-time Markov Chain such that the stability of the planar $\ppcd$
can be deduced exactly by algorithmically checking certain properties
of the reduced system. 
We study two notions of stability, namely, absolute stability and
almost sure stability. In the former, we seek to ensure that every
execution converges, while in the latter, we require that the
probability of the set of system executions that converge be $1$.
Absolute convergence ignores the probabilities associated with the
transitions, and hence, can be solved using previous results on
stability analysis of Piecewise Constant Derivative
systems~\cite{prabhakar2013abstraction},
where one checks for certain diverging transitions and cycles.
Checking almost sure convergence is much more challenging.
We show that almost sure convergence can be characterized by 
certain constraints based on the stationary distribution of the reduced system.
For this result to hold, we need mild conditions on the $\ppcd$ that
ensure the existence of this stationary distribution. 
The proof relies on several insights, including the properties of planar dynamics, and convergence 
results on infinite sequences of random variables. 

The rest of the paper is organized as follows. In section \ref{sec: related}, we discuss related works. In section \ref{sec: case}, we model motion of a planar robot with faulty angle actuator using $\ppcd$. In section \ref{sec: prelim}, we define important definitions and notations related to Markov Chains. In section \ref{sec:Stability-DTMC}, we develop algorithms for analyzing convergence of Markov Chains. We analyze stability of general and planar $\ppcd$s in section \ref{sec: PPCD}. Finally, we conclude in section \ref{sec: conc}.


\section{Related Work}\label{sec: related}
Stability is a well studied problem in classical control theory, where Lyapunov function based methods 
have been extensively developed. They have been extended to hybrid systems using multiple and 
common Lyapunov functions 
\cite{Branicky98,davrazos2001review,liberzon2003switching,van2000introduction}. However, 
constructing Lyapunov functions is computationally challenging, hence, alternate approximate 
methods have been explored. For example, in one approach the state space is divided into certain 
regions and shown that the system inevitably ends up in a certain region, thus ensuring stability 
\cite{duggirala2011abstraction,duggirala2012lyapunov,Podelski2006,Podelski2007}. Another 
approach is based on abstraction, where a simplified model (known as the abstract model) is created 
based on the original model and stability analysis on the simplified model is mapped back to the original 
one \cite{Asarin2007Hybridization,Chen2012,Prabhakar2009,Dierks2007,alur2003counter,clarke2003abstraction,prabhakar2013hybrid,prabhakar2013abstraction}.

While stability has been extensively studied in non-probabilistic setting, investigations of stability for 
probabilistic systems are limited. Sufficient conditions for stability of Stochastic Hybrid Systems via 
Lyapunov functions is discussed in the survey \cite{TEEL20142435}. Almost sure exponential stability 
\cite{cheng2012almost,cheng2018almost,do2020almost,hu2008almost} and asymptotic stability in 
distribution \cite{yuan2003asymptotic,wang2019asymptotic} for Stochastic Hybrid Systems have 
also been studied. Most of these works on probabilistic stability analysis provide approximate mehtods 
for analysis. We provide a simple class of Stochastic Hybrid Systems that have practical application in 
modeling planar robots, and an exact decidable algorithm for probabilistic stability analysis.

\section{Case Study: Planar robot with a faulty actuator}\label{sec: case}

\begin{figure}
	\centering
	\setlength\abovecaptionskip{-20pt}
	\setlength\belowcaptionskip{-10pt}
	\includegraphics[width=13cm]{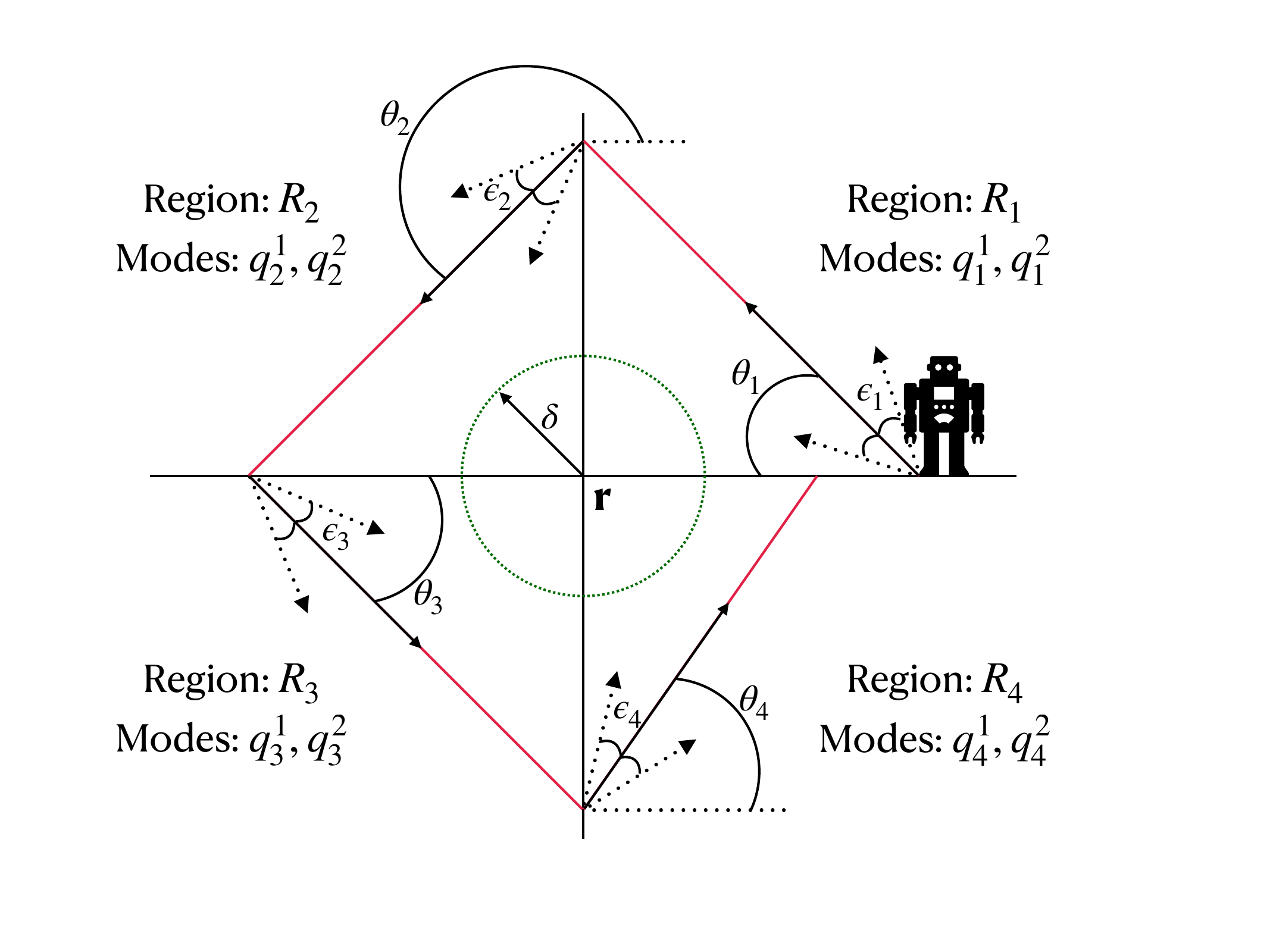}
	\caption{Motion of planar robot with faulty heading angle actuator}
	\label{planar_robot}
      \end{figure}

      Consider a robot navigating in a 2D plane at some constant speed
      $v$ as shown in Figure \ref{planar_robot}.
      The plane is divided into four regions $R_1, R_2, R_3, R_4$
      corresponding to the four quadrants, and the robot has a unique
      direction $\theta_i$ (mode of operation) in which it moves while in
      the region $R_i$, and changes its mode of operation at the
      boundary of the regions.
      Due to faulty actuator, the robot heading angle may deviate
      from $\theta_i$ by an amount $\epsilon_i$.
      We model this as probabilistically choosing one of the
      $k_i$ uniformly distanced angles $\theta_i^1, \cdots, \theta_i^{k_i}$
      in the interval $[\theta_i - \epsilon_i, \theta_i + \epsilon_i]$
      with probabilities $p_i^1, \cdots, p_i^{k_i}$, respectively.
      The whole system can be modelled as a planar $\ppcd$ with
      $\sum_{i=1}^{4} k_i$ modes, where for every $i$ and $1 \leq j
      \leq k_i$, the mode $q_i^j$ corresponds to the robot traversing
      with heading angle $\theta_i^j$ with speed $v$ in the region $R_i$.
      The mode switching is possible between $R_i$ and $R_j$ if they
      are neighbors, that is, they share a common boundary.
      For instance, we can switch between quadrants $1$ and $2$ or $4$ and $1$
      but not $1$ and $3$. 
      We can move to any mode corresponding to a neighbor $q_i^j$ with
      probability $p_i^j$.

The objective of the navigation is to reach a target point $r$ on the
2D plane arbitrarily closely. More precisely, we want to check whether
the robot reaches within a $\delta > 0$ ball around $r$ for any
arbitrarily small $\delta$.
We want to check if all executions of the robot have this property, i.e., if the planar $\ppcd$ is
absolutely stable, as well as if the probability of convergence is
$1$, i.e., the planar $\ppcd$ is almost surely stable.
      
\section{Preliminaries}\label{sec: prelim}
In this section, we will discuss important concepts related to Discrete-time Markov Chain ($\dtmc$), 
Weighted Discrete-time Markov Chain ($\wdtmc$) and convergence of $\wdtmc$.
\subsection{Discrete-time Markov Chain}
Let $\dist{\states}$ denote the set of all probability distributions on the set $\states$. Let us define 
Discrete-time Markov Chain ($\dtmc$) on the set of states $S$.
\begin{definition}[Discrete-time Markov Chain]
	The Discrete-time Markov Chain ($\dtmc$) is defined as the tuple $\calM=(\states,\tran)$ where
	\begin{itemize}
		\item $\states$ is a set of states.
		\item $\tran:\states\mapsto\dist{\states}$  is a function from the set of states $\states$ to the 
		set of all probability distributions over $\states$, $\dist{\states}$.
	\end{itemize}
\end{definition}
We use $\tran(s_1,s_2)$ to denote $\tran(s_1)(s_2)$ and $\tran^n(s_1,s_2)$ to denote the 
probability of going from $s_1$ to $s_2$ in $n$-steps.

A path of a $\dtmc$ $\calM$ is a sequence of states 
$\Path=s_1,s_2,\dots$ such that for all $i<\size{\Path}$, $\tran(s_i,s_{i+1})>0$, where 
$\size{\Path}$ is the length of the sequence. A path of length $2$ is called an edge and the set of all 
edges is denoted as $\edgeset$. The $i^{\textit{th}}$ state 
of the path $\Path$ is denoted by 
$\Path_i$ and the last state of $\Path$ is denoted as $\Path_{end}$.
$\Path[i:j]$ denotes the subsequence $\Path_i,\Path_{i+1},\dots,\Path_j$. We say $s_2$ is reachable 
from $s_1$ (denoted $s_1\leadsto s_2$) if there is a path $\Path$ on $\calM$ such that 
$\Path_1=s_1$ and $\Path_{end}=s_2$. The set of all finite paths of a $\dtmc$ $\calM$ is 
denoted as $\fpath{\calM}$ and the set of all infinite paths is denoted as $\ipath{\calM}$.

The probability of a finite path $\Path$, denoted $\tran(\Path)$, is the product of the probabilities of each of 
its edges, $\tran(\Path)\defn\prod_{i<\size{\Path}}\tran(\Path_i,\Path_{i+1})$.
The probability of $\Path$ with respect to a distribution $\rho$, denoted $\probpath{\rho}{\Path}$ is 
the product of $\tran(\Path)$ and the probability of $\Path_1$ under $\rho$, i.e.,
$\probpath{\rho}{\Path}\defn \rho(\Path_1)\cdot\tran(\Path)$.
We can associate a probability measure $\prob$ to the 
set of infinite paths $\ipath{\calM}$ of a $\dtmc$ $\calM$ using probability of the cylinder sets 
of the finite paths as discussed in \cite{Baier2008}. 
A path property $\pathprop$ is said to be \emph{almost surely} satisfied if the set of all paths having property $\pathprop$ has probability $1$, i.e., $\prob\{\Path\mid \Path\text{ has }\pathprop\}=1$.

Next we define some subclasses of $\dtmc$ and show that it has some nice convergence properties.
\begin{definition}
	[Irreducibility] A $\dtmc$ $\calM$ is called irreducible if for any 
	$s_1,s_2\in\states$, $s_1\leadsto s_2$ and $s_2\leadsto s_1$.
\end{definition}

\begin{definition}
	[Periodicity] A state $s\in\states$ in a $\dtmc$ $\calM$ is called periodic if there is a natural number $n>1$ such that,
	for any path $\Path$ starting and ending at $s$, $\size{\Path}$ is a multiple of $n$. A $\dtmc$ $\calM$ is called aperiodic if none of its states is periodic.
\end{definition}

We say a probability distribution is stationary for a $\dtmc$ $\calM$ if the next step distribution remains 
unchanged.
\begin{definition}
	[Stationary Distribution]\label{stn_dist} A distribution $\rhost\in\dist{\states}$ is 
	called the stationary distribution of $\dtmc$ $\calM$ if,
	\begin{equation*}
		\rhost(s) = \sum_{s^\prime\in\states}\rhost(s^\prime)\tran(s^\prime,s),\quad \forall s\in\states.
	\end{equation*}
\end{definition}

For finite, irreducible $\dtmc$, the stationary distribution is unique. The following theorem guarantees existence of limiting distribution for finite,
irreducible and aperiodic $\dtmc$ and associates it with the stationary distribution of the $\dtmc$ (see \cite{ROSS2010191}).
\begin{theorem}\label{exist_stationary}
	For a finite, irreducible and aperiodic $\dtmc$ $\lim_{n\rightarrow\infty} \tran^n(s_1,s_2)$ 
	 exists for all $s_1,s_2\in\states$ and $\lim_{n\rightarrow\infty} 
	\tran^n(s_1,s_2) = \rhost(s_2)$ where $\rhost\in\dist{\states}$ is the unique stationary distribution 
	of 	$\calM$.
\end{theorem}
Note that, $\tran^n(s_1,s_2)$ does not depend on $s_1$ as $n\rightarrow\infty$.

\subsection{Weighted Discrete-time Markov Chain}
Let us now define Weighted Discrete-time Markov Chain ($\wdtmc$) that extend $\dtmc$ with weighted edges. Basically, a $\wdtmc$ can be observed as a Markov Reward Process where rewards are associated to individual transitions rather than nodes.

\begin{definition}
	[Weighted $\dtmc$] The weighted $\dtmc$ ($\wdtmc$) $\calMw=(\states,\tran,\weight)$ is a tuple such 
	that $(\states,\tran)$ is a $\dtmc$ and $\weight:\edgeset\mapsto\real$ is a weight function where 
	$\edgeset$ is the set of all possible edges of $\calMw$.
\end{definition}
We also define disjoint union of two $\wdtmc$ $\calMw^1$ and $\calMw^2$ as a $\wdtmc$ $\calMw^1\disunion\calMw^2$ whose states and edges are disjoint unions of states and edges of $\calMw^1$ and $\calMw^2$ respectively.
With the weight function $\weight$ defined, it is possible to associate weights to individual paths of 
$\calMw$.
\begin{definition}
	[Weight of a path]\label{wt_path_def} The weight of a path $\Path$ of $\wdtmc$ $\calMw$, denoted 
	$\weight(\Path)$, is defined as,
	\begin{equation*}
		\weight(\Path)\defn\sum_{i<\size{\Path}}\weight(\Path_i,\Path_{i+1})
	\end{equation*}
\end{definition}

For $\Path\in\ipath{\calMw}$, the quantity $\lim_{n\rightarrow\infty}\sum_{i=1}^{n}\weight(\Path_i,\Path_{i+1})$ is denoted by $\weight(\Path[1:\infty])$. It is easy to observe that, $\weight(\Path)=\weight(\Path[1:\infty])$.
A simple path is a path without state repetition and a simple cycle is a path where only the starting and 
the ending states are same. 
We use the notation $\spath(\calMw)$ for the set of all simple paths and the notation $\scyl(\calMw)$ for the set of all 
simple cycles of a $\wdtmc$ $\calMw$.

\subsection{Convergence of Weighted Discrete-time Markov Chain}
Let us define the notions of absolute and probabilistic convergence of $\wdtmc$. A $\wdtmc$ is said to 
be absolutely convergent if the weight of every infinite path diverges to $-\infty$.
\begin{definition}
	[Absolute Convergence of $\wdtmc$]\label{ac_wdtmc}	A $\wdtmc$ $\calMw$ is said to be absolutely 
	convergent if for all infinite path $\Path\in\ipath{\calMw}$, $\weight(\Path)$ diverges to $-\infty$, i.e.,
	\begin{equation*}
		\weight(\Path[1:\infty]) = -\infty.
	\end{equation*}
\end{definition}
Further, a $\wdtmc$ is said to be almost surely convergent if the weight of an infinite path diverges to $-\infty$ with probability $1$.
\begin{definition}
	[Almost Sure Convergence of $\wdtmc$]\label{asc_wdtmc} We say that a $\wdtmc$
	$\calMw$ is almost 
	surely convergent if for any path $\Path$ of $\calMw$, 
	$\weight(\Path)$ diverges to $-\infty$ 
	with probability $1$. In other words,
	\[\prob\left\{\Path\in\ipath{\calMw}: \weight(\Path[1:\infty]) = -\infty\right\}=1.\]
\end{definition}

\begin{remark}
Let us explain the reason behind defining such a strange notion of convergence. 
For reasons that will be clarified later, we actually want to check for an infinite path $\Path$ of $\calMw$, if the product of weights of the edges converge to $0$, i.e., $\lim_{n\rightarrow\infty}\prod_{i=1}^n\weight(\Path_i,\Path_{i+1}) = 0$, provided $0<\weight(\Path_i,\Path_{i+1})<\infty$ for all $i\in\nat$. This condition is equivalent to $\lim_{n\rightarrow\infty}\sum_{i=1}^n\log(\weight(\Path_i,\Path_{i+1})) = -\infty$. Hence for convenience, we consider $\log$ of original weights as weights of individual edges, and check if sum of weights of edges of an infinite path diverge to $-\infty$.
\end{remark}

\subsection{Probabilistic Bisimulation}
Probabilistic bisimulation~\cite{Baier2008} on a $\wdtmc$ is an equivalence relation on its set of states such that probabilities of corresponding edges agree for two related states.
\begin{definition}[Probabilistic Bisimulation]
    A probabilistic bisimulation on a $\wdtmc$ $\calMw$ is an equivalence relation $\sim$ on $\states$ such that for any $s_1,s_2\in\states$ with $s_1\sim s_2$, $\tran(s_1,T)=\tran(s_2,T)$ for each equivalence class $T$ of $\sim$.
\end{definition}
Note that, $\tran(s,T)=\sum_{t\in T}\tran(s,t)$ for $s\in \states$.
Let us now use probabilistic bisimulation to relate infinite paths of a $\wdtmc$.
\begin{definition}
    [Bisimulation-Equivalent Paths] Given a probabilistic bisimulation $\sim$ on a $\wdtmc$ $\calMw$, two infinite paths $\pi=\pi_1,\pi_2,\dots$ and $\tilde{\pi}=\tilde{\pi}_1,\tilde{\pi}_2,\dots$ are said to be bisimulation equivalent, denoted $\pi\sim\tilde{\pi}$, if they are statewise related by $\sim$, i.e.,
    \[\pi\sim\tilde{\pi}\text{ iff }\pi_i\sim\tilde{\pi}_i\text{ for all }i\geq 1\]
\end{definition}
A set of infinite paths is $\sim$ bisimulation-closed for some probabilistic bisimulation $\sim$, if for any path in the set, all its bisimulation-equivalent paths are also in the set. In other words, $\Pi\subseteq\ipath{\calMw}$ is $\sim$ bisimulation-closed if for any $\pi\in\Pi$ and any $\tilde{\pi}\sim\pi$, $\tilde{\pi}\in\Pi$. Let us denote by $\prob_s(\Pi)$ the set of all paths in $\Pi$ that start from $s\in\states$. The following lemma~\cite{Baier2008} equates the probability of two sets of paths that start from $\sim$ related states and are subset of the same $\sim$ bisimulation-closed set.

\begin{lemma}\label{lem: bisim_prob}
Let $\sim$ be a probabilistic bisimulation on a $\wdtmc$ $\calMw$.
For all states $s_1,s_2$ of $\calMw$, $s_1\sim s_2$ implies $\prob_{s_1}(\Pi)=\prob_{s_2}(\Pi)$, for all $\sim$ bisimulation-closed events $\Pi\subseteq\ipath{\calMw}$.
\end{lemma}

\subsection{Polyhedral Sets}
We denote the set of all polyhedral subsets of $\real^n$ by $\poly{n}$. The facets of a polyhedral 
subset $A$ are the largest polyhedral subsets of the boundary of $A$. We denote the boundary of a 
polyhedral subset $A$ by $\boundary{A}$ and the set of all facets of $A$ by $\face(A)$. We say a 
polyhedral subset $P$ is positive scaling invariant if for all $x\in P$ and $\alpha>0$, $\alpha x\in P$.
\section{Analyzing Convergence of Weighted Discrete-time Markov Chains}
\label{sec:Stability-DTMC}

In this section, we discuss necessary and sufficient conditions for absolute and almost 
sure convergence of $\wdtmc$. For our analysis, we will assume all paths of the $\wdtmc$ start from a 
single state called the \emph{initialization point} (denoted $\initst$) of the $\wdtmc$. In other words we 
restrict our attention to the set of paths 
$\pathset{}{\prime}{}\defn\{\Path\in\ipath{\calMw}\mid\Path_1=\initst\}$.
Consequently, we consider only those edges $\edgeset^\prime =
\pathset{}{\prime}{}\intersect\edgeset$, which are reachable from $\initst$. We abuse notation and use 
$\pathset{}{}{}$ for $\pathset{}{\prime}{}$ and $\edgeset$ for $\edgeset^\prime$ for the rest of the section.

\subsection{Analyzing absolute convergence of Weighted $\dtmc$}

Here we provide a necessary and sufficient condition for analyzing absolute convergence of a 
$\wdtmc$. We begin with the following proposition (proved in the Appendix) which states that for any finite path 
$\Path\in\fpath{\calMw}$,
we can get one simple path and a set of simple cycles such that their total weight equals the 
weight of $\Path$.

\begin{proposition}\label{path_decomp_sc_sp}
	For any finite path $\Path$ of $\calMw$ there exist a simple path $\Path_s\in\spath(\calMw)$ and a set of 
	simple cycles $\scyl_\Path\subseteq\scyl(\calMw)$ such that 
	$\weight(\Path)=\weight(\Path_s)+\sum_{\cycle\in\scyl_\Path}\weight(\cycle)$.
\end{proposition}

We use Proposition \ref{path_decomp_sc_sp} to prove the following main theorem which states that, a 
$\wdtmc$ is absolutely convergent iff there is no edge of infinite weight and no cycle of weight greater or equal to $0$
reachable from the initial point.
\begin{theorem}\label{abs_conv}
	The $\wdtmc$ $\calMw$ is absolutely convergent iff,
	\begin{enumerate}
		\item There does not exist an edge $e\in\edgeset$ reachable from $\initst$ such that 
		$\weight(e)=\infty$.
		\item For any simple cycle $\cycle$ reachable from $\initst$, 
		$W(\cycle)< 0$.
	\end{enumerate}
\end{theorem}
\begin{proof}
		($\Rightarrow$) To show that the conditions $1$ and $2$ are necessary, we have to prove that if 
		either of them is negated then $\calMw$ is not absolutely convergent.
	If condition $1$ is false then there is an edge $e = (s_1,s_2)$ with $\weight(s_1,s_2)=\infty$ such 
	that for some finite path $\Path$ starting from $\initst$, $\Path_{\size{\Path}-1}=s_1$ and 
	$\Path_{\size{\Path}}=s_2$. But that implies 
	$\weight(\Path)=\sum_{i=1}^{\size{\Path}-1}\weight(\Path_i,\Path_{i+1})=\infty$. So for any infinite path $\Path^\prime$ with prefix $\Path$, 
	$\weight(\Path^\prime)=\infty$. Thus $\calMw$ is not absolutely convergent. 
	On the other hand if we suppose condition $2$ is false then there is a simple cycle $\cycle\in\scyl(\calMw)$ 
	with $\weight(\cycle)\geq 0$ such that for some finite path $\Path$ starting from $\initst$, there exists 
	an index $j$ such that $\cycle=\Path[j:\size{\Path}]$. Now we can easily construct the following 
	infinite path $\infpath=\Path\cdot\cycle\cdot\cycle\dots$ by concatenating $\cycle$ infinite times to 
	$\Path$. Clearly, $\infpath$ starts at $\initst$ since $\Path$ starts at $\initst$ and 
	$\weight(\infpath)=\weight(\Path)+\sum_{n\in\nat}\weight(\cycle)\geq \weight(\Path)$. Since for any finite path $\Path$, $\weight(\Path)$ is also finite, $\weight(\infpath)$ is bounded below by some finite quantity and cannot diverge to $-\infty$. Thus, $\calMw$ is not absolutely convergent.
	
	($\Leftarrow$) Conversely, suppose both conditions $1$ and $2$ hold. 
	Now, let $\Path$ be an arbitrary infinite path starting from $\initst$ and $\Path[1:i]$ be its finite prefix of length $i\in\nat$. By Proposition \ref{path_decomp_sc_sp}, there exist a simple path ${\Path[1:i]}_s$ and a set of simple cycles $\scyl_{\Path[1:i]}$ such that $\weight(\Path[1:i]) = \weight({\Path[1:i]}_s) +  \sum_{\cycle\in\scyl_{\Path[1:i]}}\weight(\cycle)$. Now, for any $i\in\nat$, $\weight({\Path[1:i]}_s)$ is at most $\sum_{(s_1,s_2)\in\edgeset}\max\{\weight(s_1,s_2)\mid (s_1,s_2)\in\edgeset\}<\infty$. Also, $\scyl_{\Path[1:i]}$ is a set of simple cycles where each cycle has weight at most $\max_{\cycle\in\scyl(\calMw)}\weight(\cycle)< 0$ (here we abuse notation and denote the set of all simple cycles reachable from $\initst$ as $\scyl(\calMw)$). Thus, for all $K\in\real$, there exists $i\in\nat$ such that
	\begin{align*}
	    &\sum_{(s_1,s_2)\in\edgeset}\max\{\weight(s_1,s_2)\mid (s_1,s_2)\in\edgeset\} + \sum_{\cycle\in\scyl_{\Path[1:i]}}\weight(\cycle)< K\\
	    &\Rightarrow \weight({\Path[1:i]}_s) +  \sum_{\cycle\in\scyl_{\Path[1:i]}}\weight(\cycle) < K\\
	    &\Rightarrow \weight(\Path[1:i]) < K.
	\end{align*}
	But this implies $\weight(\Path)=\lim_{i\rightarrow\infty}\weight(\Path[1:i])=-\infty$ for any infinite path $\Path$ starting from $\initst$, i.e., $\calMw$ is absolutely convergent.\hfill$\qed$
\end{proof}


\subsection{Analyzing almost sure convergence of Weighted $\dtmc$}
In this subsection, we will provide a necessary and sufficient condition for almost sure convergence of 
a $\wdtmc$. We assume a $\wdtmc$ $\calMw$ is finite, irreducible and aperiodic and thus has the limiting 
distribution equal to its stationary distribution $\rhost$ (Theorem \ref{exist_stationary}).

Given a $\wdtmc$ $\calMw$, we begin by defining random variables $\{X^e_j\mid e\in\edgeset;j\in\nat\}$ on the set of infinite paths $\ipath{\calMw}$, that captures the information of whether an edge $e\in\edgeset$ appears on the $j^{\textit{th}}$ step of an infinite path $\Path$. More precisely,
\begin{equation*}
	X^e_j(\Path) = 
	\begin{cases}
		1 \text{ if }(\Path_j,\Path_{j+1})=e\\
		0 \text{ else}.
	\end{cases}
\end{equation*}
Note that for some $e\in\edgeset$ and $\Path\in\ipath{\calMw}$, $\sum_{j=1}^n X^e_j(\Path)$ gives the number of times $e$ appears on $\Path[1:n+1]$.
Now, the following lemma (proved in Appendix) gives that, for any edge $e\in\edgeset$, the average of $\{X^e_j\mid j\in\nat\}$ almost surely converges to $\probpath{\rhost}{e}$, which is the probability of $e$ with respect to the stationary distribution $\rhost$.

\begin{lemma}\label{lem: as_avg_prob}
    For any edge $e\in\edgeset$ of a $\wdtmc$ $\calMw$,
    \[
        \prob\left\{\Path\in\ipath{\calMw} : \lim_{n\rightarrow\infty}\frac{\sum_{j=1}^n X_j^e(\Path)}{n}=\probpath{\rhost}{e}\right\} = 1.
    \]
\end{lemma}
Next, we define partial average weight upto $n$ for an infinite path $\Path$ as
\begin{equation*}
	\frac{(\wld)_n}{n} := \frac{\sum_{i=1}^{n}\weight(\Path_i,\Path_{i+1})}{n},
\end{equation*}
and note that,
\begin{align}\label{eq: sn_relate_X}
	\frac{(\wld)_n}{n}&=\frac{\sum_{e\in\edgeset}(\text{\# times }e\text{ appears on 
		}\Path[1:n+1])\cdot\weight(e)}{n}\nonumber\\
	&=\frac{\sum_{e\in\edgeset}\left(\sum_{j=1}^n 
	X^e_j(\Path)\right)\cdot\weight(e)}{n}
\end{align}
We now state the main lemma of this subsection which essentially states that, the average weight of an infinite path almost surely converges to a quantity that depends only on the weights and probabilities of the edges.
\begin{lemma}\label{lim_lasso}
    For a $\wdtmc$ $\calMw$,
    \[
        \prob\left\{\Path\in\ipath{\calMw} : \lim_{n\rightarrow\infty}\frac{(\wld)_n}{n} = \sum_{e\in\edgeset}\probpath{\rhost}{e}\weight(e)\right\} = 1.
    \]
\end{lemma}
\begin{proof}
    We have already established that,
\begin{align*}
    &\frac{(\wld)_n}{n} = \frac{\sum_{e\in\edgeset}\left(\sum_{j=1}^n 
	X^e_j(\Path)\right)\cdot\weight(e)}{n}\quad[\text{Equation }\ref{eq: sn_relate_X}]\\
    \text{Thus, }&\lim_{n\rightarrow\infty}\frac{(\wld)_n}{n} =
    \lim_{n\rightarrow\infty}\frac{\sum_{e\in\edgeset}\left(\sum_{j=1}^n 
	X^e_j(\Path)\right)\cdot\weight(e)}{n}\\
	\Rightarrow & \lim_{n\rightarrow\infty}\frac{(\wld)_n}{n} = \sum_{e\in\edgeset}\probpath{\rhost}{e}\cdot\weight(e)\text{ almost surely}\quad[\text{by Lemma }\ref{lem: as_avg_prob}]
\end{align*}\hfill$\qed$
\end{proof}
We say $\sum_{e\in\edgeset}\probpath{\rhost}{e}\weight(e)$ is the effective weight of the $\wdtmc$ 
$\calMw$ and denote it as $\weight_\edgeset$.
The main theorem basically states that a $\wdtmc$ is almost surely convergent iff its effective weight is strictly less than $0$.
\begin{theorem}\label{thm_as_conv}
	A $\wdtmc$ $\calMw$ is almost surely convergent iff 
	$\weight_\edgeset < 0$, where $\weight_\edgeset=\sum_{e\in\edgeset}\probpath{\rhost}{e}\weight(e)$ is the effective weight of $\calMw$.
\end{theorem}
\begin{proof}
    Observe that, weight of an infinite path $\Path$, $\weight(\Path)$, can be written as $\lim_{n\rightarrow\infty}n\cdot\left((\wld)_n/n\right)$, where $\left((\wld)_n/n\right)$ is the partial average weight upto $n$ for the infinite path $\Path$. Since,
	\begin{align*}
	&\lim_{n\rightarrow\infty}\left(\frac{(\wld)_n}{n}\right) = \sum_{e\in\edgeset}\probpath{\rhost}{e}\weight(e)\text{ almost surely}\quad[\text{by Lemma }\ref{lim_lasso}]\\
	\Rightarrow &\weight(\Path) =  \lim_{n\rightarrow\infty}n\cdot\left(\frac{(\wld)_n}{n}\right)
	= \lim_{n\rightarrow\infty}n\cdot\left(\sum_{e\in\edgeset}\probpath{\rhost}{e}\weight(e)\right)\text{ almost surely}
	\end{align*}
	Thus, $\weight(\Path)$ diverges to $-\infty$ almost surely if and only if $\sum_{e\in\edgeset}\probpath{\rhost}{e}\weight(e) < 0$.
	In other words, $\calMw$ is almost surely convergent iff $\weight_\edgeset< 0$.\hfill$\qed$
\end{proof}

\subsection{Computability}\label{sec: comp}
Based on Theorems \ref{abs_conv} and \ref{thm_as_conv} we present two algorithms here for 
checking absolute and almost sure convergence of a $\wdtmc$. 
For the first algorithm, assuming the $\wdtmc$ is finite, we 
first check for
existence of an infinite weight edge by Breadth First Search (BFS) 
\cite{kleinberg2006algorithm} 
and then for a cycle with non-negative weight using a 
variant of the Bellman-Ford algorithm \cite{kleinberg2006algorithm}. 
If neither of them is found then 
the $\wdtmc$ is deemed absolutely convergent by Theorem \ref{abs_conv}. 
Since BFS takes time linear to the size of its input and Bellman-Ford takes time quadratic to the size of 
its input, the time complexity of this algorithm is $O(\size{\states}^2)$, where 
$\states$ is the set of states of $\calMw$.

For the second algorithm, assuming the $\wdtmc$ is finite, irreducible and aperiodic, 
existence of an infinite weight edge is checked by Breadth First Search (BFS). If such an edge exists then the 
$\wdtmc$ is deemed not almost surely convergent (by Theorem \ref{thm_as_conv}).
Otherwise, the 
stationary distribution $\rhost$ of the 
$\wdtmc$ is calculated by solving a set of linear equations mentioned in Definition \ref{stn_dist}. The value 
$\sum_{e\in\edgeset}\probpath{\rhost}{e}\weight(e)$ is then calculated (where $\edgeset$ is the set of 
transitions of the $\wdtmc$) and compared to $0$. The $\wdtmc$ is deemed almost surely convergent 
only if $\sum_{e\in\edgeset}\probpath{\rhost}{e}\weight(e)< 0$. 
Since BFS takes time linear to its input size and solving a set of linear equations takes time at most 
cubic in the number of variables, the time complexity of this algorithm is 
$O(\size{\states}^3)$, where $\states$ is the set of states of $\calMw$.

\section{Probabilistic Piecewise Constant Derivative Systems}\label{sec: PPCD}
In this section, we present the details of the Probabilistic Piecewise
Constant Derivative Systems ($\ppcd$) and provide a characterization of
absolute and almost sure stability by a reduction to that of DTMCs.

\subsection{Formal Definition of $\ppcd$}
We model $\ppcd$s as consisting of a discrete set of modes, each
associated with an invariant and probabilistic transitions between
modes that are enabled at the boundaries of the invariants.
\begin{definition}
	[$\ppcd$]\label{def-ppcd}
	The Probabilistic Piecewise Constant Derivative System ($\ppcd$) is defined as the tuple 
	$\calH\defn(\loc,\state,\inv,\flow,\edges)$ where
	\begin{itemize}
		\item $\loc$ is the set of discrete locations,
		\item $\state = {\real}^n$ is the continuous state space for
		some $n\in\nat$,
		\item $\inv: \loc\rightarrow\poly{n}$ is the invariant
                  function which assigns a positive scaling invariant polyhedral subset of the state space to each location $q\in\loc$,
		\item $\flow: \loc\rightarrow \state$ is the Flow function which 
		assigns a flow vector, say $\flow(q)\in\state$, to 
		each location $q\in \loc$,
		\item $\edges \subseteq
                  \loc \times(\union_{q\in\loc}\face(\inv(q)))\times\dist{Q}$
                  is the probabilistic edge relation such that $(q, f,
                  \rho)\in\edges$ where for every $(q, f)$, there is a
                  at most one $\rho$ such that $(q, f, \rho)  \in
                  \edges$ and $f\in\face(\inv(q))$.
                  $f$ is called a \emph{Guard} of the location $q$.
	\end{itemize}
\end{definition}
      
Next, we discuss the semantics of the $\ppcd$.
An execution starts from a location $q_0\in\loc$ 
and some continuous state $x_0\in\state$ and evolves continuously for
some time $T$ according to the dynamics of $q_0$ until it reaches a
facet $f_0$ of the invariant of $q_0$.
Then a probabilistic discrete transition is taken if there is an edge
$(q_0, f_0, \rho_0)$ and the state $q_0$ is probabilistically changed
to $q_1$ with probability $\rho_0(q_1)$.
The execution (tree) continues with alternating continuous and
discrete transitions.

Formally, for any two continuous states $x_1, x_2\in\state$ and $q \in
\loc$, we say that there is a \emph{continuous transition} from $x_1$
to $x_2$ with respect to  $q$ if $x_1,x_2\in\inv(q)$, there exists
$T\geq 0$ such that $x_2=x_1+\flow(q)\cdot T$, $x_1 + \flow(q) \cdot t
\not \in \boundary{\inv(q_0)}$ for any $0 \leq t < T$ and $x_2 \in
\boundary{\inv(q_0)}$.
We note that there is a unique continuous transition from any state
$(q, x)$ since it requires the state to evolve until it reaches the
boundary for the first time, which corresponds to a unique time of 
evolution $T$.
Further, if for all $t\geq 0$, $x_1+\flow(q)\cdot t\in\inv(q)$ then we
say $x_1$ has an infinite edge with respect to $q$.
For two locations $q_1, q_2\in\loc$, we say there is a \emph{discrete
transition} from $q_1$ to $q_2$ with probability $p$ via $\rho\in\dist{\loc}$ and
$f\in\face(q_1)$ if  $f\subseteq\inv(q_2)$, $(q_1,f,\rho)\in\edges$
and $p = \rho(q_2)$.

We capture the semantics of a $\ppcd$ using a $\wdtmc$, wherein we combine a
continuous transition and a discrete transition to represent a
probabilistic transition of the $\dtmc$. In addition, to reason
about convergence, we also need to capture the relative distance of
the states from the equilibrium point, which is captured using edge weights.
Let us fix $0$ as the equilibrium point for the rest of the
section.
The weight on a transition from $(q_1, x_1)$ to $(q_2, x_2)$ captures
the logarithm of the relative distance of $x_1$ and $x_2$ from $0$,
that is, it is $\left(\norm{x_2}{\infty}/\norm{x_1}{\infty}\right)$,
where $\norm{x}{\infty}$ captures the distance of state $x$ from $0$.

\begin{definition}
	[Semantics of $\ppcd$]
	Given a $\ppcd$ $\calH$, we can construct the $\wdtmc$
        $\semantics{\calH}\defn(\states_\calH,\tran_\calH,\weight_\calH)$
        where, 
	\begin{itemize}
		\item $\states_\calH=\loc\times\state$
		\item $\tran_\calH$ and $\weight_\calH$ are defined as
                  follows for any $(q_1, x_1)$ and $(q_2, x_2)$:
		\begin{itemize}
			\item If there is a continuous transition from
                          $x_1$ to $x_2$ with respect to $q_1$  and there is a 
			discrete transition from $q_1$ to $q_2$ with
                        probability $p$ via some $\rho\in\dist{\loc}$
                        and $f\in\face(q_1)$, and $x_2\in f$, then
                        $\tran_\calH((q_1,x_1),(q_2,x_2)) = p$ and
                        $\weight_\calH((q_1,x_1),(q_2,x_2))=\log \left(\norm{x_2}{\infty}/\norm{x_1}{\infty}\right)$
			\item If $x_1$ has an infinite edge with
                          respect to $q_1$, then
                          $\tran_\calH((q_1,x_1),(q_2,x_2))= 1$ if
                          $(q_1,x_1)=(q_2,x_2)$ and $0$, otherwise,
                          and $\weight_\calH((q_1,x_1),(q_1,x_1))=\infty$.
                      \item
                        Otherwise, $\tran_\calH((q_1,x_1),(q_2,x_2))$
                        $= \weight_\calH((q_1,x_1),$ $(q_2,x_2))= 0$.

                      \end{itemize}

	\end{itemize}
      \end{definition}

Since all executions of the $\ppcd$ $\calH$ start from location $q_0$ and state $x_0$, we consider only those paths of the semantics $\semantics{\calH}$ which start from $(q_0,x_0)$ and denote them as $\ipath{\semantics{\calH}}$. We say a path $\Path = (q_0,x_0),(q_1,x_1),\dots$ converges to $0$ if norm of the corresponding state sequence $\norm{x_0}{\infty},\norm{x_1}{\infty},\dots$ converges to $0$. Stability of a $\ppcd$ $\calH$ is defined in terms of convergence of paths of its semantics $\semantics{\calH}$ as follows,

\begin{definition}
    [Stability of $\ppcd$]\label{def:stab_ppcd}
    A $\ppcd$ $\calH$ is called absolutely stable if every path of  $\semantics{\calH}$ converges to $0$. Analogously, $\calH$ is called almost surely stable if any path of $\semantics{\calH}$ converges to $0$ with probability $1$, i.e.,
    \begin{equation*}
        \prob\left\{\Path\in\ipath{\semantics{\calH}}: \Path\text{ converges to }0\right\}=1.
    \end{equation*}
\end{definition}

We now characterize stability of a $\ppcd$ $\calH$ in terms of its semantics $\semantics{\calH}$. Basically we state that, $\calH$ is absolutely (almost surely) stable iff $\semantics{\calH}$ is absolutely (almost surely) convergent.

\begin{theorem}[Characterization of Stability]\label{thm: char_stable}
A $\ppcd$ $\calH$ is absolutely stable iff its semantics $\semantics{\calH}$ is absolutely convergent and it is almost surely stable iff $\semantics{\calH}$ is almost surely convergent.
\end{theorem}

\begin{proof}
Note that, a path $\Path$ of $\semantics{\calH}$ converges to $0$ iff $\weight(\Path)$ diverges to $-\infty$. To observe this, let $\Path = (q_0,x_0),(q_1,x_1),\dots$. Then, $\norm{x_0}{\infty},\norm{x_1}{\infty},\dots$ converge to $0$ iff,
\begin{align*}
    &\lim_{n\rightarrow\infty}\frac{\norm{x_n}{\infty}}{\norm{x_0}{\infty}} = 0
    \quad[\text{since }\norm{x_0}{\infty}\neq 0]\\
    \iff & \lim_{n\rightarrow\infty} \frac{\norm{x_1}{\infty}}{\norm{x_0}{\infty}}\cdot
    \frac{\norm{x_2}{\infty}}{\norm{x_1}{\infty}}\cdot\cdot\cdot
    \frac{\norm{x_n}{\infty}}{\norm{x_{n-1}}{\infty}} = 0\\
    &
    \begin{aligned}
    \   \   \   \   \   \   \   \   \ 
    [\text{since }\norm{x_i}{\infty}\neq 0\text{ for all } i=1,\dots,n-1\text{ if }\Path \text{ is infinite}]
    \end{aligned}\\
    \iff & \lim_{n\rightarrow\infty}
    \log\left(\frac{\norm{x_1}{\infty}}{\norm{x_0}{\infty}}\cdot
    \frac{\norm{x_2}{\infty}}{\norm{x_1}{\infty}}\cdot\cdot\cdot
    \frac{\norm{x_n}{\infty}}{\norm{x_{n-1}}{\infty}}\right) = -\infty\\
    \iff & \lim_{n\rightarrow\infty}
    \log\left(\frac{\norm{x_1}{\infty}}{\norm{x_0}{\infty}}\right)+
    \log\left(\frac{\norm{x_2}{\infty}}{\norm{x_1}{\infty}}\right)+\cdot\cdot\cdot
    \log\left(\frac{\norm{x_n}{\infty}}{\norm{x_{n-1}}{\infty}}\right) = -\infty\\
    \iff & \lim_{n\rightarrow\infty}\weight(\Path[1:n]) = -\infty
\end{align*}
Thus, every infinite path of $\semantics{\calH}$ converges to $0$ iff weight of every infinite path diverges to $-\infty$ and the 
set of infinite paths of $\semantics{\calH}$ converging to $0$ has probability $1$ iff the set of infinite paths having weight diverging to $-\infty$ has probability $1$.
In other words, $\calH$ is absolutely (almost surely) stable iff $\semantics{\calH}$ is absolutely (almost surely) convergent.\hfill$\qed$
\end{proof}

\subsection{Stability of Planar $\ppcd$}
\begin{figure}
	\centering
	\setlength\abovecaptionskip{-20pt}
	\setlength\belowcaptionskip{-10pt}
	\includegraphics[width=12cm]{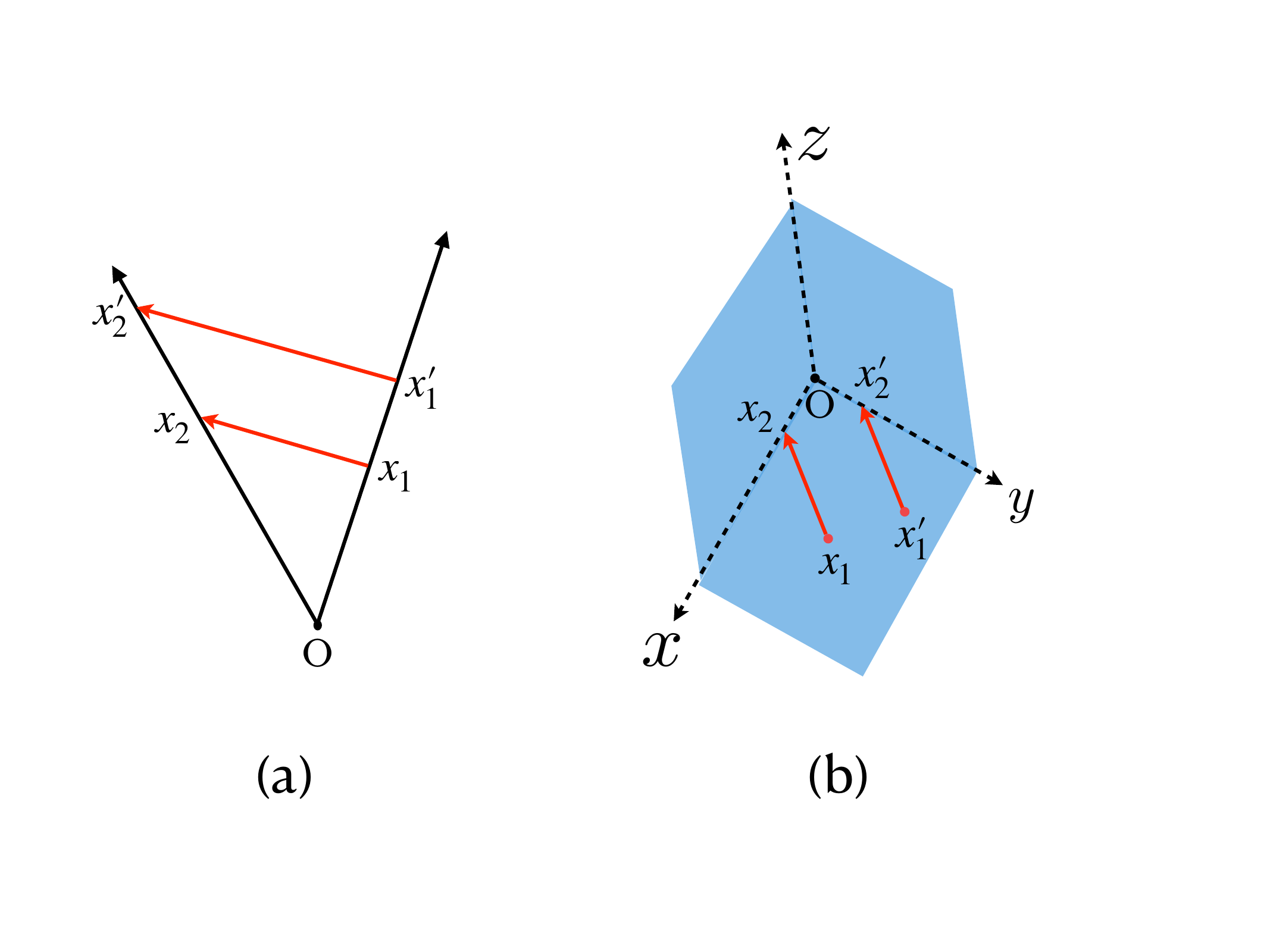}
	\caption{(a) In $\real^2$, continuous transition with constant rate starting from any point in a facet leads to a unique point in a unique facet. (b) In $\real^3$, even with constant rate, continuous transitions starting from different points in the same facet may end up in different facets.}
	\label{fig: unique_2D}
\end{figure}
In general, semantics of a $\ppcd$ has infinite number of states and thus the algorithms developed in section \ref{sec: comp} cannot be applied to decide absolute (almost sure) convergence of the semantics. However, if the continuous state space of a $\ppcd$ $\calH$ is $\real^2$, then we can reduce $\semantics{\calH}$ to a finite $\wdtmc$ that provides an exact characterization of $\semantics{\calH}$. A $\ppcd$ with $\state=\real^2$ is called a planar $\ppcd$. 
Since for each location $q$, $\inv(q)$ is positively scaled, the facets of $\inv(q)$ are rays emanating from origin. 
Given constant flow for each location $q$, a continuous transition starting at a point of some facet $f_1\in \union_{q\in\loc}\face(\inv(q))$ ends up at a unique point of a unique facet $f_2\in\union_{q\in\loc}\face(\inv(q))$. 
This property is not observed if the continuous state space is of three or higher dimensions (Figure \ref{fig: unique_2D}).
Also, if two continuous transitions start from different points $x_1,x_1^\prime$ of the same facet $f_1$, they end up in unique points $x_2,x_2^\prime$ (respectively) of a unique facet $f_2$ such that $\norm{x_2}{\infty}/\norm{x_1}{\infty}=\norm{x_2^\prime}{\infty}/\norm{x_1^\prime}{\infty}$. This gives us the following lemma (proved in Appendix),

\begin{lemma}\label{obs-init-independent}
Let $e  = ((q_1,x_1), (q_2,x_2))$, $e' = ((q_1,x'_1), (q_2,x'_2))$ be two edges of $\semantics{\calH}$ (where $\calH$ is a planar $\ppcd$) such
that, $\tran_\calH(e), \tran_\calH(e') > 0$, and $x_1,x_1^\prime\in f_1$ where  $f_1\in\bigunion_{q\in\loc}\face(\inv(q))$.
Then $\tran_\calH(e) = \tran_\calH(e')$ and $\weight_\calH(e) = \weight_\calH(e')$.
\end{lemma}

For the rest of the section, we will assume all paths of the semantics $\semantics{\calH}$ of a planar $\ppcd$ $\calH$ start at $(q_0,x_0)$ and $x_0\in f_0$, where $f_0$ is a facet in $\union_{q\in\loc}\face(\inv(q))$.

We now define the quotient of a planar $\ppcd$ $\calH$, which is a finite $\wdtmc$ having the same convergence properties as $\semantics{\calH}$. Here we consider the set of states as $\loc\times\union_{q\in\loc}\face(\inv(q))$ instead of $\loc\times\state$ and use Lemma \ref{obs-init-independent} to define the probabilistic edges and their weights.

\begin{definition}
	[Quotient of $\ppcd$] Let $\calH$ be a planar $\ppcd$ and $\semantics{\calH}$ be its 
	semantics. We define the $\wdtmc$ 
	$\reduce{\calH}=(\reduce{\states},\reduce{\tran},\reduce{\weight})$ as follows,
	\begin{itemize}
		\item $\reduce{\states} = \loc\times\bigunion_{q\in\loc}\face(\inv(q))$
		\item $\reduce{\tran}((q_1,f_1),(q_2,f_2))
                  =\tran_\calH((q_1,x_1),(q_2,x_2))$ for some $x_1 \in
                  f_1$ and $x_2 \in f_2$ such that $\tran_\calH ((q_1,
                  x_1), (q_2, x_2)) > 0$, and $0$ otherwise.
                  \item
                  $\reduce{\weight}((q_1,f_1),(q_2,f_2))= \weight_\calH((q_1,x_1),(q_2,x_2))$ for some $x_1 \in
                  f_1$ and $x_2 \in f_2$ such that $\tran_\calH ((q_1,
                  x_1), (q_2, x_2)) > 0$, and $0$ otherwise.
		\end{itemize}
\end{definition}

The above definition is well-defined, that is, the choice of $x_1$ and
$x_2$ do not matter due to Lemma \ref{obs-init-independent}.

We will eventually prove that a planar $\ppcd$ $\calH$ is absolutely (almost surely) stable if and only if its quotient $\wdtmc$ $\reduce{\calH}$ is absolutely (almost surely) convergent. First, let us show that for every infinite path $\Path$ of $\semantics{\calH}$, there is a path $\pi$ in $\reduce{\calH}$ with same weight and vice versa. 
\begin{lemma}
	[Conservation of weight]\label{prop-weight-conserve} For every infinite path $\Path$ 
	of $\semantics{\calH}$, there is a path 
	$\pi$ in $\reduce{\calH}$ such that $\weight(\Path)=\weight(\pi)$ and vice versa.
\end{lemma}
\begin{proof}
($\Rightarrow$) Let $\Path=\Path_1,\Path_2,\dots$ be an infinite path of $\semantics{\calH}$. By assumption, 
	$\Path_i\in f_i$ where $f_i\in\bigunion_{q\in\loc}\face(q)$ is a facet, for each $i\in\nat$. Suppose for 
	each $i$, $\Path_i = (q_i,x_i)$. Since for each $i$, there is an edge between $(q_i,x_i)$ and 
	$(q_{i+1},x_{i+1})$ in $\semantics{\calH}$, there should be an edge between $(q_i,f_i)$ and 
	$(q_{i+1},f_{i+1})$ in $\reduce{\calH}$. Using 
	Lemma \ref{obs-init-independent} we can conclude that for all $i$,
	$\weight((q_i,x_i),(q_{i+1},x_{i+1}))=\weight((q_i,f_i),(q_{i+1},f_{i+1}))$. Thus we can construct the 
	infinite path $\pi=((q_1,f_1),(q_2,f_2),\dots)$ such that $\weight(\pi)=\weight(\Path)$.
	
($\Leftarrow$) To prove the converse, we show by induction that for any $n\in\nat$ if there is a path $\pi$ of length $n$ in 
	$\reduce{\calH}$ then there is a path $\Path$ of length $n$ in $\semantics{\calH}$ with same 
	weight as $\pi$.\\
	\textbf{Base case}: Suppose $((q_1,f_1),(q_2,f_2))$ is an edge of $\reduce{\calH}$. Then there 
	exist $x_1\in f_1$ and $x_2\in f_2$ such that $x_2=x_1+\flow(q_1)\cdot t$, for some $t\geq 0$, i.e., 
	there is an edge 
	between $(q_1,x_1)$ and $(q_2,x_2)$ in $\semantics{\calH}$. Also by Lemma 
	\ref{obs-init-independent}, $\weight((q_1,x_1),(q_2,x_2))=\weight((q_1,f_1),(q_2,f_2))$. Hence base 
	case is proved.
	
	Now suppose $((q_1,f_1),\dots,(q_n,f_n),(q_{n+1},f_{n+1}))$ is a path of $\reduce{\calH}$ and by 
	induction hypothesis we have a path $((q_1,x_1),\dots,(q_n,x_n))$ in $\semantics{\calH}$ such that 
	$\weight((q_1,f_1),\dots,(q_n,f_n))=\weight((q_1,x_1),\dots,(q_n,x_n))$. Since there is an edge 
	between $(q_n,f_n)$ and $(q_{n+1},f_{n+1})$, there exist $x^\prime_n\in f_n$ and 
	$x^\prime_{n+1}\in f_{n+1}$ such that 
	\begin{equation}\label{st_exist}
		x^\prime_{n+1} = x^\prime_n + \flow(q_n)\cdot t
	\end{equation}
	for some $t\geq 0$. Since $\state=\real^2$, $f_n$ and $f_{n+1}$ are rays. By Equation 
	\ref{st_exist}, there is a straight line of slope $\flow(q_n)$ that intersects both of them. But then 
	any straight line with slope $\flow(q_n)$ intersecting $f_n$ will also intersect $f_{n+1}$, in fact, if 
	we take the straight line with slope 
	$\flow(q_n)$ passing through $x_n$, it will intersect $f_{n+1}$. That means there exists $t\geq 0$ 
	and $x_{n+1}\in f_{n+1}$ such that $x_{n+1} = x_n+\flow(q_n)\cdot t$. This is because for $t$ to 
	be negative, $f_n$ and $f_{n+1}$ must intersect and $x_n$ and $x^\prime_n$ must lie on opposite 
	sides of this intersection point on $f_n$. But this is impossible since $f_n$ and $f_{n+1}$ intersect 
	only at $0$ and both of them get terminated at $0$. Thus there exist $x_{n+1}\in f_{n+1}$ such 
	that $((q_n,x_n),(q_{n+1},x_{n+1}))$ is an edge of $\semantics{\calH}$. By Lemma 
	\ref{obs-init-independent}, $\weight((q_n,x_n),(q_{n+1},x_{n+1})) = 
	\weight((q_n,f_n),(q_{n+1},f_{n+1}))$. Hence our claim is proved for all $n\in\nat$, i.e., it holds for 
	infinite paths of $\reduce{\calH}$ as well.\hfill$\qed$
\end{proof}

Using Lemma \ref{prop-weight-conserve}, we now prove the main theorem which states that a $\ppcd$ is absolutely (almost surely) stable if and only if its quotient $\wdtmc$ is absolutely (almost surely) stable.

\begin{theorem}\label{thm: quotient-ppcd-relate}
    A planar $\ppcd$ $\calH$ is absolutely (almost surely) stable iff its quotient $\wdtmc$ $\reduce{\calH}$ is absolutely (almost surely) convergent.
\end{theorem}

\begin{proof}
A $\ppcd$ $\calH$ is absolutely stable iff $\semantics{\calH}$ is absolutely convergent (Theorem \ref{thm: char_stable}). By Lemma \ref{prop-weight-conserve}, it is easy to observe that every infinite path of $\semantics{\calH}$ diverge to $-\infty$ if and only if every infinite path of $\reduce{\calH}$ diverge to $-\infty$. Thus, we can conclude that $\calH$ is absolutely stable if and only if $\reduce{\calH}$ is absolutely stable.

On the other hand, a $\ppcd$ $\calH$ is almost surely stable iff  $\semantics{\calH}$ is almost surely convergent (Theorem \ref{thm: char_stable}). 
Let us show that $\semantics{\calH}$ is almost surely convergent iff $\reduce{\calH}$ is almost surely convergent.
Since we have assumed that all paths of $\semantics{\calH}$ start from $(q_0,x_0)$, all paths of $\reduce{\calH}$ will start from $(q_0,f_0)$, where $f_0$ is the facet containing $x_0$. Let us define the equivalence relation $\sim$ on the set of states of the $\wdtmc$ $\semantics{\calH}\disunion\reduce{\calH}$ as,
\begin{align*}
    &(q_i,x_i)\sim(q_j,x_j) \text{ if } q_i=q_j\text{ and }x_i,x_j\text{ belong to the same facet}\\
    &(q_i,x_i)\sim(q_j,f_j)\text{ if } q_i=q_j\text{ and }x_i\in f_j\\
    &(q_i,f_i)\sim(q_j,f_j)\text{ if } q_i=q_j\text{ and }f_i=f_j,
\end{align*}
where $q_i,q_j\in\loc$, $x_i,x_j\in\state$ and $f_i,f_j\in\union_{q\in\loc}\face(\inv(q))$. Note that, the set of equivalence classes of $\sim$ is given by $\{(q,f)\mid q\in\loc, f\in\face(\inv(q))\}$. Now by Lemma \ref{obs-init-independent}, we can easily deduce that $\sim$ is a probabilistic bisimulation on $\semantics{\calH}\disunion\reduce{\calH}$. Observe that, the set \[\Pi = \{\pi\in\ipath{\semantics{\calH}\disunion\reduce{\calH}} :  \weight(\pi[1:\infty]) = -\infty\}\]
is $\sim$ bisimulation-closed. To see this, take any $\pi\in\Pi$ and $\tilde{\pi}\sim\pi$. By Lemma \ref{obs-init-independent}, $\weight(\pi_i,\pi_{i+1})=\weight(\tilde{\pi}_i,\tilde{\pi}_{i+1})$ for all $i$. Thus, $\weight(\tilde{\pi}[1:\infty])=-\infty$ as well, i.e., $\tilde{\pi}\in\Pi$. Now, we have $\prob_{(q_0,x_0)}(\Pi) = \prob_{(q_0,f_0)}(\Pi)$ as a direct consequence of Lemma \ref{lem: bisim_prob}, i.e.,
\begin{align*}
    &\prob\{\Path\in\ipath{\semantics{\calH}}\mid\Path_1 = (q_0,x_0)\text{ and } \weight(\Path) \text{ diverges to }-\infty\}\\
    =& \prob\{\pi\in\ipath{\reduce{\calH}}\mid\pi_1 = (q_0,f_0)\text{ and } \weight(\pi) \text{ diverges to }-\infty\}.
\end{align*}
Hence, $\prob\{\Path\in\ipath{\semantics{\calH}}\mid\Path_1 = (q_0,x_0)\text{ and } \weight(\Path) \text{ diverges to }-\infty\}=1$ if and only if $\prob\{\pi\in\ipath{\reduce{\calH}}\mid\pi_1 = (q_0,f_0)\text{ and } \weight(\pi) \text{ diverges to }-\infty\}=1$, i.e., $\semantics{\calH}$ is almost surely convergent iff $\reduce{\calH}$ is almost surely convergent.
Thus, $\calH$ is almost surely stable iff $\reduce{\calH}$ is almost surely convergent. \hfill$\qed$
\end{proof}

Since $\reduce{\calH}$ is finite, we can use the algorithms developed in section \ref{sec: comp} to decide its absolute (almost sure) convergence. This in turn decides absolute (almost sure) stability of $\calH$ by Theorem \ref{thm: quotient-ppcd-relate}.
\section{Conclusion}\label{sec: conc}
In this paper, we showed the decidability of absolute and almost sure
convergence of Planar Probabilistic Piecewise Constant 
Derivative Systems ($\ppcd$), that are a practically useful subclass of
stochastic hybrid systems and can model motion of planar robots with
faulty actuators.
We give a computable characterization of absolute and almost sure
convergence through a reduction to a finite state $\dtmc$.
In the future, we plan to extend these ideas to analyze higher
dimensions $\ppcd$ and SHS with more complex dynamics.
In particular, the idea of reduction can be applied to higher
dimensional $\ppcd$ but we will need to extend our analysis to a Markov
Decision Process that will appear as the reduced system.

%
%
%
 \bibliographystyle{splncs04}
 \bibliography{Spandan}

\begin{thebibliography}{10}
\providecommand{\url}[1]{\texttt{#1}}
\providecommand{\urlprefix}{URL }
\providecommand{\doi}[1]{https://doi.org/#1}

\bibitem{alur2003counter}
Alur, R., Dang, T., Ivan{\v{c}}i{\'c}, F.: Counter-example guided predicate
  abstraction of hybrid systems. In: International Conference on Tools and
  Algorithms for the Construction and Analysis of Systems. pp. 208--223.
  Springer (2003)

\bibitem{Asarin2007Hybridization}
Asarin, E., Dang, T., Girard, A.: Hybridization methods for the analysis of
  nonlinear systems. Acta Informatica  \textbf{43},  451--476 (2007).
  \doi{10.1007/s00236-006-0035-7}

\bibitem{Baier2008}
Baier, C., Katoen, J.P.: Principles of Model Checking (Representation and Mind
  Series). The MIT Press (2008)

\bibitem{Branicky98}
Branicky, M.: Multiple lyapunov functions and other analysis tools for switched
  and hybrid systems. IEEE Transactions on Automatic Control  \textbf{43}(4),
  475--482 (1998). \doi{10.1109/9.664150}

\bibitem{Chen2012}
Chen, X., Ábrahám, E., Sankaranarayanan, S.: Taylor model flowpipe
  construction for non-linear hybrid systems. In: 2012 IEEE 33rd Real-Time
  Systems Symposium. pp. 183--192 (2012). \doi{10.1109/RTSS.2012.70}

\bibitem{cheng2012almost}
Cheng, P., Deng, F.: Almost sure exponential stability of linear impulsive
  stochastic differential systems. In: Proceedings of the 31st Chinese Control
  Conference. pp. 1553--1557. IEEE (2012)

\bibitem{cheng2018almost}
Cheng, P., Deng, F., Yao, F.: Almost sure exponential stability and stochastic
  stabilization of stochastic differential systems with impulsive effects.
  Nonlinear Analysis: Hybrid Systems  \textbf{30},  106--117 (2018)

\bibitem{clarke2003abstraction}
Clarke, E., Fehnker, A., Han, Z., Krogh, B., Ouaknine, J., Stursberg, O.,
  Theobald, M.: Abstraction and counterexample-guided refinement in model
  checking of hybrid systems. International journal of foundations of computer
  science  \textbf{14}(04),  583--604 (2003)

\bibitem{davrazos2001review}
Davrazos, G., Koussoulas, N.: A review of stability results for switched and
  hybrid systems. In: Mediterranean Conference on Control and Automation.
  Citeseer (2001)

\bibitem{Dierks2007}
Dierks, H., Kupferschmid, S., Larsen, K.G.: Automatic abstraction refinement
  for timed automata. In: Raskin, J.F., Thiagarajan, P.S. (eds.) Formal
  Modeling and Analysis of Timed Systems. pp. 114--129. Springer Berlin
  Heidelberg, Berlin, Heidelberg (2007)

\bibitem{do2020almost}
Do, K.D., Nguyen, H.: Almost sure exponential stability of dynamical systems
  driven by l{\'e}vy processes and its application to control design for
  magnetic bearings. International Journal of Control  \textbf{93}(3),
  599--610 (2020)

\bibitem{duggirala2011abstraction}
Duggirala, P.S., Mitra, S.: Abstraction refinement for stability. In: 2011
  IEEE/ACM Second International Conference on Cyber-Physical Systems. pp.
  22--31. IEEE (2011)

\bibitem{duggirala2012lyapunov}
Duggirala, P.S., Mitra, S.: Lyapunov abstractions for inevitability of hybrid
  systems. In: Proceedings of the 15th ACM international conference on Hybrid
  Systems: Computation and Control. pp. 115--124 (2012)

\bibitem{HENZINGER199894}
Henzinger, T.A., Kopke, P.W., Puri, A., Varaiya, P.: What's decidable about
  hybrid automata? Journal of Computer and System Sciences  \textbf{57}(1),
  94--124 (1998). \doi{https://doi.org/10.1006/jcss.1998.1581},
  \url{https://www.sciencedirect.com/science/article/pii/S0022000098915811}

\bibitem{hu2008almost}
Hu, L., Mao, X.: Almost sure exponential stabilisation of stochastic systems by
  state-feedback control. Automatica  \textbf{44}(2),  465--471 (2008)

\bibitem{kleinberg2006algorithm}
Kleinberg, J., Tardos, E.: Algorithm Design. Addison-Wesley (2006)

\bibitem{lal2018hierarchical}
Lal, R., Prabhakar, P.: Hierarchical abstractions for reachability analysis of
  probabilistic hybrid systems. In: 2018 56th Annual Allerton Conference on
  Communication, Control, and Computing (Allerton). pp. 848--855. IEEE (2018)

\bibitem{lal2019counterexample}
Lal, R., Prabhakar, P.: Counterexample guided abstraction refinement for
  polyhedral probabilistic hybrid systems. ACM Transactions on Embedded
  Computing Systems (TECS)  \textbf{18}(5s),  1--23 (2019)

\bibitem{liberzon2003switching}
Liberzon, D.: Switching in systems and control. Springer Science \& Business
  Media (2003)

\bibitem{Podelski2006}
Podelski, A., Wagner, S.: Model checking of hybrid systems: From reachability
  towards stability. In: Hespanha, J.P., Tiwari, A. (eds.) Hybrid Systems:
  Computation and Control. pp. 507--521. Springer Berlin Heidelberg, Berlin,
  Heidelberg (2006)

\bibitem{Podelski2007}
Podelski, A., Wagner, S.: A sound and complete proof rule for region stability
  of hybrid systems. In: Bemporad, A., Bicchi, A., Buttazzo, G. (eds.) Hybrid
  Systems: Computation and Control. pp. 750--753. Springer Berlin Heidelberg,
  Berlin, Heidelberg (2007)

\bibitem{prabhakar2013hybrid}
Prabhakar, P., Duggirala, S., Mitra, S., Viswanathan, M.: Hybrid automata-based
  cegar for rectangular hybrid automata. In: In http://www. its. caltech.
  edu/pavithra/Papers/rtss2012tr. pdf. Citeseer (2013)

\bibitem{prabhakar2013abstraction}
Prabhakar, P., Soto, M.G.: Abstraction based model-checking of stability of
  hybrid systems. In: International Conference on Computer Aided Verification.
  pp. 280--295. Springer (2013)

\bibitem{prabhakar2013decidability}
Prabhakar, P., Viswanathan, M.: On the decidability of stability of hybrid
  systems. In: Proceedings of the 16th international conference on Hybrid
  systems: computation and control. pp. 53--62 (2013)

\bibitem{Prabhakar2009}
Prabhakar, P., Vladimerou, V., Viswanathan, M., Dullerud, G.E.: Verifying
  tolerant systems using polynomial approximations. In: 2009 30th IEEE
  Real-Time Systems Symposium. pp. 181--190 (2009). \doi{10.1109/RTSS.2009.28}

\bibitem{prajna2004safety}
Prajna, S., Jadbabaie, A.: Safety verification of hybrid systems using barrier
  certificates. In: International Workshop on Hybrid Systems: Computation and
  Control. pp. 477--492. Springer (2004)

\bibitem{ROSS2010191}
Ross, S.M.: Chapter 4 - markov chains. In: Ross, S.M. (ed.) Introduction to
  Probability Models (Tenth Edition), pp. 191--290. Academic Press, Boston,
  tenth edn. (2010). \doi{https://doi.org/10.1016/B978-0-12-375686-2.00009-1}

\bibitem{rutten2004mathematical}
Rutten, J.J., Kwiatkowska, M., Norman, G., Parker, D.: Mathematical techniques
  for analyzing concurrent and probabilistic systems. No.~23, American
  Mathematical Soc. (2004)

\bibitem{TEEL20142435}
Teel, A.R., Subbaraman, A., Sferlazza, A.: Stability analysis for stochastic
  hybrid systems: A survey. Automatica  \textbf{50}(10),  2435--2456 (2014).
  \doi{https://doi.org/10.1016/j.automatica.2014.08.006},
  \url{https://www.sciencedirect.com/science/article/pii/S0005109814003070}

\bibitem{van2000introduction}
Van Der~Schaft, A.J., Schumacher, J.M.: An introduction to hybrid dynamical
  systems, vol.~251. Springer London (2000)

\bibitem{wang2019asymptotic}
Wang, B., Zhu, Q.: Asymptotic stability in distribution of stochastic systems
  with semi-markovian switching. International Journal of Control
  \textbf{92}(6),  1314--1324 (2019)

\bibitem{yuan2003asymptotic}
Yuan, C., Mao, X.: Asymptotic stability in distribution of stochastic
  differential equations with markovian switching. Stochastic processes and
  their applications  \textbf{103}(2),  277--291 (2003)

\end{thebibliography}
%
\section*{Appendix}
\subsection*{Proof of Proposition \ref{path_decomp_sc_sp}}
Here we provide a detailed proof of Proposition \ref{path_decomp_sc_sp} which states that,\newline
\emph{
    For any finite path $\Path$ of $\calMw$ there exist a simple path $\Path_s\in\spath(\calMw)$ and a set of 
	simple cycles $\scyl_\Path\subseteq\scyl(\calMw)$ such that 
	$\weight(\Path)=\weight(\Path_s)+\sum_{\cycle\in\scyl_\Path}\weight(\cycle)$.
}
\begin{proof}
	We traverse $\Path$ and whenever a cycle $\cycle$ is encountered, remove its edges from $\Path$ 
	and add the cycle to the set $\scyl_\Path$. This process is repeated until $\scyl_\Path$ 
	contains only simple cycles and the remaining edges of $\Path$ form a simple path 
	$\Path_s=\Path-(\union\{\cycle\mid\cycle\in\scyl_\Path\})$. Let $\edgeset^{\Path_s}$ denote the set 
	of 
	edges of $\Path_s$ and for each $\cycle\in\scyl_\Path$, $\edgeset^\cycle$ denote the set of edges 
	of 
	$\cycle$. Clearly, $\{\edgeset^{\Path_s}\}\union\{\edgeset^\cycle\mid\cycle\in\scyl_\Path\}$ is a 
	partition of 
	the set of edges of $\Path$. Thus 	
	$\weight(\Path)=\weight(\Path_s)+\sum_{\cycle\in\scyl_\Path}\weight(\cycle)$. Hence, our claim 
	is proved.\hfill$\qed$
\end{proof}

\subsection*{Algorithms from Section \ref{sec: comp}}
Based on the discussions of section \ref{sec: comp}, we provide pseudocodes for algorithms for checking absolute (almost sure) convergence of a finite (finite, irreducible and aperiodic) $\wdtmc$.
\begin{algorithm}[H]
\begin{algorithmic}[1]
    \renewcommand{\algorithmicrequire}{\textbf{Input:}}
    \renewcommand{\algorithmicensure}{\textbf{Output:}}
	\Require{A $\wdtmc$ $\calMw \defn (\states,\tran,\weight)$}
	\Ensure Yes/No
	\State Convert $\calMw$ to a weighted graph $\graph=(\vertx,\gredge,\ewt^\prime)$ where,\newline
	$\vertx=\states$, 
	$\gredge = \{(s_1,s_2)\in \states\times\states\mid \tran(s_1,s_2)>0\}$,\newline 
	and $\ewt^\prime:\gredge\rightarrow\real$ defined as $\ewt^\prime(e) \defn -\weight(e)$
	\State Run BFS on $\graph$ to check existence of edge with weight $-\infty$
	\If{(edge with $-\infty$ weight exists)}
		\State Return No
	\EndIf
	\State Run Bellman-Ford algorithm on $\graph$
	\If{(cycle with negative weight is found)}
	    \State Return No
	\Else
	    \State Let $d:\vertx\rightarrow \real_{\geq 0}$ define the shortest distance of each $v\in\vertx$ from $\initst$
	    \State Mark in $E$ all edges $(u,v)$ such that $d(v)=d(u)+\ewt^\prime(u,v)$
	    \State Delete from $\graph$ all unmarked edges
	    \State Run DFS on $\graph$ (with unmarked edges deleted) to check for a cycle
	    \If{(a cycle is found)}
	        \State Return No
	    \Else
		    \State Return Yes
		\EndIf
	\EndIf
\end{algorithmic}
\caption{Checking absolute convergence of $\wdtmc$}\label{alg_abs_conv}
\end{algorithm}
\begin{algorithm}[H]
    \begin{algorithmic}[1]
    \renewcommand{\algorithmicrequire}{\textbf{Input:}}
    \renewcommand{\algorithmicensure}{\textbf{Output:}}
	\Require A $\wdtmc$ $\calMw \defn (\states,\tran,\weight)$
	\Ensure Yes/No
	\State Convert $\calMw$ to a weighted graph $\graph=(\vertx,\gredge,\ewt^\prime)$ where,\newline 
	$\vertx=\states$, 
	$\gredge = \{(s_1,s_2)\in \states\times\states\mid \tran(s_1,s_2)>0\}$,\newline
	and $\ewt^\prime:\gredge\rightarrow\real$ defined as $\ewt^\prime(e) \defn \weight(e)$
	\State Run BFS on $\graph$ to check existence of edge with weight $\infty$
	\If{(edge with $\infty$ weight exists)}
		\State Return No
	\EndIf
	\State Calculate stationary distribution $\rhost$ of $\calMw$ by solving the set of linear equations,
	\begin{align*}
		&\rhost(s) = \sum_{s^\prime\in\states}\rhost(s^\prime)\tran(s^\prime,s),\quad \forall s\in\states\\
		&\sum_{s\in\states}\rhost(s) = 1
	\end{align*}
	\State $asWeight \gets 0$\;
	\For{$e\in \gredge$}
		\State $asWeight = asWeight + \probpath{\rhost}{e}\ewt^\prime(e)$
	\EndFor
	\If{$asWeight<0$}
		\State Return Yes
	\Else
		\State Return No
	\EndIf
	\end{algorithmic}
	\caption{Checking almost sure convergence of $\wdtmc$}\label{alg_as_conv}
\end{algorithm}

\subsection*{Proof of Lemma \ref{lem: as_avg_prob}}
We prove Lemma \ref{lem: as_avg_prob} here which essentially states that,\newline
\emph{
For any edge $e\in\edgeset$ of a $\wdtmc$ $\calMw$,
    \[
        \prob\left\{\Path\in\ipath{\calMw} : \lim_{n\rightarrow\infty}\frac{\sum_{j=1}^n X_j^e}{n}=\probpath{\rhost}{e}\right\} = 1,
    \]
}
\begin{proof}
    Construct the $\dtmc$ $\calM' = (\states',\tran')$ from $\calMw$, where $\states' = \states\union \edgeset$ and for each $e = (s,s')\in\edgeset$, $(s,e),(e,s')\in\edgeset'$ with $\tran'(s,e)=\tran(s,s')$ and $\tran'(e,s')=1$ ($\edgeset'$ is the set of edges of $\calM'$). Note that, there is a one to one correspondence between $\ipath{\calMw}$ and $\ipath{\calM'}$, where each edge $e=(s,s')$ in $\Path\in\ipath{\calMw}$ is replaced by consecutive edges $(s,e)$ and $(e,s')$ in the corresponding path $\Path'\in\ipath{\calM'}$. Thus, $(\Path_j,\Path_{j+1})=e$ if and only if $\Path'(2j)=e$, where $\Path'$ is the corresponding path of $\Path$. Now, let us define random variables $\{Y_j^x\mid x\in\states'; j\in\nat\}$ as,
    \begin{equation*}
        Y^x_j = 
        \begin{cases}
            1 \text{ if } \Path'_j = x\\
            0 \text{ else}
        \end{cases}
    \end{equation*}
    for $\Path'\in\ipath{\calM'}$.
    Then, it is easy to observe that, $\sum_{j=1}^n X^e_j = \sum_{j=1}^{2n}Y^e_{2j}$. Note that, $\calM'$ is finite and irreducible. Hence, by strong law of large numbers for any $x\in\states'$ \cite{ROSS2010191},
    \begin{equation*}
        \lim_{n\rightarrow\infty} \frac{\sum_{j=1}^{2n} Y^x_j}{2n} = {\rhost}'(x)\text{ almost surely},
    \end{equation*}
    where ${\rhost}'$ is the stationary distribution of $\calM'$. Since for any $x\in\states'$,
    \begin{align}\label{eq: avg_limit_as}
        &\lim_{n\rightarrow\infty} \frac{\sum_{j=1}^{2n} Y^x_{2j}}{2n} = 
        \lim_{n\rightarrow\infty} \frac{\sum_{j=1}^{2n} Y^x_j}{2n}\nonumber\\
        \text{Thus, } & \lim_{n\rightarrow\infty}\frac{\sum_{j=1}^n X_j^e}{n} = 2\left(\lim_{n\rightarrow\infty} \frac{\sum_{j=1}^{2n} Y^e_{2j}}{2n}\right) = 2{\rhost}'(e)\text{ almost surely}.
    \end{align}
  
    Consider $\rho:\states'\rightarrow[0,1]$ as
    \begin{equation*}
        \rho(x) = 
        \begin{cases}
            \frac{\rhost(x)}{2} \text{ if }x\in\states\\
            \frac{\tran(x)\rhost(s)}{2} \text{ if }x=(s,s')\in\edgeset.
        \end{cases}
    \end{equation*}
    where $\rhost$ is the stationary distribution of $\calMw$.
    Let us observe that, $\sum_{x\in\states'}\rho(x) = \sum_{s\in\states}\rhost(s)/2 + \sum_{s\in\states}\sum_{(s,s')\in\edgeset}\rhost(s)\tran(s,s')/2 = 1$, i.e., $\rho$ is a probability distribution. 
    
    Note that, for any $x\in\states$,
    \begin{align*}
        \sum_{x'\in\states'}\rho(x')\tran'(x',x)
        &=\sum\{\rho(e)\tran'(e,x) : e = (s',x)\in\edgeset\}\\
        &=\sum\left\{\frac{\tran(e)\rhost(x)}{2} : e = (s',x)\in\edgeset\right\}\\
        &=\frac{\rhost(x)}{2} = \rho(x).
    \end{align*}
    
    And for any $x=(s,s')\in\edgeset$,
    \begin{equation*}
        \sum_{x'\in\states'}\rho(x')\tran'(x',x)
        =\rho(s)\tran'(s,x)
        =\frac{\rhost(s)}{2}\cdot\tran(x)=\rho(x).
    \end{equation*}
    Thus, for all $x\in\states'$, $\rho(x) = \sum_{x'\in\states'}\rho(x')\tran'(x',x)$, i.e., 
    $\rho$ is a stationary distribution for $\calM'$. Since $\calM'$ is finite and irreducible, it has a unique stationary distribution. Thus, $\rho={\rhost}'$, which ultimately provides for any $e=(s,s')\in\edgeset$,
    \begin{align*}
    \lim_{n\rightarrow\infty}\frac{\sum_{j=1}^n X_j^e}{n} &= 2\left(\frac{\tran(e)\rhost(s)}{2}\right)\text{ almost surely}\quad[\text{by Equation } \ref{eq: avg_limit_as}]\\
    &= \rhost(s)\tran(e)\text{ almost surely}\\
    &= \probpath{\rhost}{e}\text{ almost surely},
    \end{align*}
    This proves Lemma \ref{lem: as_avg_prob}.\hfill$\qed$
\end{proof}

\subsection*{Proof of Lemma \ref{obs-init-independent}}
We prove Lemma \ref{obs-init-independent} here which states the following,\newline
\emph{
Let $e  = ((q_1,x_1), (q_2,x_2))$, $e' = ((q_1,x'_1), (q_2,x'_2))$ be two edges of $\semantics{\calH}$ (where $\calH$ is a planar $\ppcd$) such
that, $\tran_\calH(e), \tran_\calH(e') > 0$, and $x_1,x_1^\prime\in f_1$ where  $f_1\in\bigunion_{q\in\loc}\face(\inv(q))$.
Then $\tran_\calH(e) = \tran_\calH(e')$ and $\weight_\calH(e) = \weight_\calH(e')$.
}
\begin{proof}
Since continuous state space of $\calH$ is $\real^2$, there is a unique facet $f_2$ for $f_1$ such that $x_2,x_2^\prime\in f_2$ (assuming $\weight_\calH(e),\weight_\calH(e')\neq\infty$). Now, since $\tran_\calH(e)$ 
	and $\tran_\calH(e^\prime)$ depend only on $q_1$ and $f_2$, $\tran_\calH(e)=\tran_\calH(e^\prime)$.
	Since any facet is a ray emanating from the origin, it can be depicted by the formula $y=kx$, where $k\in\real$.
	Let $x_1=(x_1[1],x_1[2])$ and $x_2=(x_2[1],x_2[2])$. By property of 
	$\ppcd$, $x_2 = x_1+\flow(q_1)\cdot T$ for some $T\geq0$. Thus,
	\begin{equation}\label{eq1}
		(x_2[1],x_2[2])
		= (x_1[1],x_1[2])+(\flow(q_1)[1]),\flow(q_1)[2])T
	\end{equation}
	Let $f_1:y=k_1x$ and $f_2:y=k_2x$. So,
	\begin{align}
		x_2[2] &= k_2\cdot x_2[1]\label{eq2}\\
		x_1[2] &= k_1\cdot x_1[1]\label{eq3}
	\end{align}
	Using equations \ref{eq1},\ref{eq2},\ref{eq3} we can write $x_2[1]=c\cdot x_1[1]$ where $c$ 
	depends on $k_1$, $k_2$, $\flow(q_1)[1]$ and $\flow(q_1)[2]$. Thus 
	$\frac{\norm{x_2}{\infty}}{\norm{x_1}{\infty}}$ can also be written in terms of $k_1$, $k_2$, 
	$\flow(q_1)[1]$ and $\flow(q_1)[2]$ since $\frac{\norm{x_2}{\infty}}{\norm{x_1}{\infty}}$ is equal to 
	either $\size{x_2[2]}/\size{x_1[2]}$ or $\size{x_2[2]}/\size{x_1[1]}$ or $\size{x_2[1]}/\size{x_1[2]}$ or 
	$\size{x_2[1]}/\size{x_1[1]}$ and $x_1$ and $x_2$ dependent terms on numerator and denomenator 
	always cancel off each other.
	Same is true for $e'$ as well. Thus, $\weight_\calH(e),\weight_\calH(e')$ depend only on $q$, $f_1$ and $f_2$ and not on the points $x_1,x_1^\prime,x_2,x_2^\prime$. Hence, they must be equal.\hfill$\qed$
\end{proof}

\end{document}